\documentclass[11pt]{article}
\usepackage{mathptmx}
\usepackage{epsfig,amsfonts,latexsym,amssymb,shadow,amsmath,wrapfig}
\usepackage{epstopdf}
\usepackage{verbatim}
\usepackage{mathtools}
\usepackage{graphicx,color,multirow,colortbl,hhline,array,url,fancyhdr,setspace,enumerate}
\usepackage{subcaption}

\usepackage[normalem]{ulem}
\usepackage[margin=1in]{geometry}
\usepackage{enumitem}
\usepackage{hyperref}
\hypersetup{hidelinks} 
\usepackage{amsmath,bm}
\usepackage{mathtools}
\usepackage{amsfonts}
\usepackage{amssymb}
\usepackage{float}
\usepackage[ruled,vlined,linesnumbered]{algorithm2e}
\usepackage{booktabs}
\usepackage{tabularx}

\usepackage{threeparttable}
\usepackage{bm}
\usepackage{xcolor}
\usepackage{caption}
\usepackage{multirow}
\usepackage[superscript,biblabel,nomove]{cite}
\usepackage{jabbrv}
\usepackage{lineno}

\newcommand{\name}{FMAE} 

\usepackage[hang,flushmargin,symbol]{footmisc} 
\usepackage{bbding}
\usepackage{pifont}
\usepackage[misc]{ifsym} 
\author{}
\date{}
\begin{document}
\title{Multitask Battery Management with Flexible Pretraining}

\author{%
  Hong Lu\textsuperscript{1,\dag},~
  Jiali Chen\textsuperscript{2,\dag},~
  Jingzhao Zhang\textsuperscript{1,\dag},~ \\
  Guannan He\textsuperscript{3~*},~
  Xuebing Han\textsuperscript{2~*},~
  Minggao Ouyang\textsuperscript{2~*}\\
}

    \footnotetext{1~~IIIS, Tsinghua University; \ Shanghai Qi Zhi Institute}
    \footnotetext{2~~State Key Laboratory of Intelligent Green Vehicle and Mobility, School of Vehicle and Mobility, Tsinghua University.}
	\footnotetext{3~~School of Advanced Manufacturing and Robotics,\ National Engineering Laboratory for Big Data Analysis and Applications, Peking University.}
	\footnotetext{\dag~~These authors contributed equally: Hong Lu, Jiali Chen, Jingzhao Zhang, }
	\footnotetext{*~~Email: gnhe@pku.edu.cn, hanxuebing@tsinghua.edu.cn, ouymg@tsinghua.edu.cn}

\maketitle 

\begin{abstract}
    Industrial-scale battery management involves various types of tasks, such as estimation, prediction, and system-level diagnostics. Each task employs distinct data across temporal scales, sensor resolutions, and data channels. Building task-specific methods requires a great deal of data and engineering effort, which limits the scalability of intelligent battery management. Here we present the Flexible Masked Autoencoder (FMAE), a flexible pretraining framework that can learn with missing battery data channels and capture inter-correlations across data snippets. FMAE learns unified battery representations from heterogeneous data and can be adopted by different tasks with minimal data and engineering efforts. Experimentally, FMAE consistently outperforms all task-specific methods across five battery management tasks with eleven battery datasets. On remaining life prediction tasks, FMAE uses 50 times less inference data while maintaining state-of-the-art results. Moreover, when real-world data lack certain information, such as system voltage, FMAE can still be applied with marginal performance impact, achieving comparable results with the best hand-crafted features. FMAE demonstrates a practical route to a flexible, data-efficient model that simplifies real-world multi-task management of dynamical systems.

\end{abstract}

\newpage

    \section{Introduction}
    Lithium-ion batteries (LiB) are prevalently applied in transportation and energy systems\cite{ncs2021-battery_for_Energy_storage, NE2021-train, NC2023-carbonNeutrality-EV, NREE2024-carbonNeutrality-EV, NE2017-carbonNeutrality-BESS}, and have been considered vital means to achieve carbon neutrality\cite{NREE2024-carbonNeutrality-EV, NE2017-carbonNeutrality-BESS, NC2024-carbonNeutrality}. The reliable operations of these LiB systems entail efficient battery management\cite{NRCT2025-effBM, ncs2024_digital_twin_battery_life_forecast}, which is rather challenging due to the non-linearity of electrochemical mechanisms and complex system dynamics of packed cells\cite{npj2024-nonLinear, CE2024-complexSystem}. 
    
    Existing battery management studies have explored the estimation of state of charge (SOC)\cite{AE2025-SOC, eTrans2024-SOC, JPS2025-SOC}, capacity\cite{AE2018-capacity, NC2024-capacity, etran2024-capacity-dataset}, internal resistance (IR)\cite{JPS2023-IR, EnAI2021-IR, eTrans2025-IR}, the prediction of remaining useful life (RUL)\cite{NE2019-RUL-dataset, NMI2025-RUL-BliNet}, and the anomaly detection\cite{NC2023-dyad-dataset, eTrans2020-fault, etrans2022-fault, NC2025-fault} for battery systems. Recent advances on data-driven battery management discovered important features and observations by analyzing hundreds of cell measures in various laboratory environments \cite{NMI2020-SohPipeline, NC2022-voltageRelaxation-dataset, NC-2023-SOH-DNNswarm}. These works greatly improve the accuracy of state-of-health estimation for battery cells. Later works \cite{NC2023-dyad-dataset, NC2025-MultiModality, NE2024-houseBESS-dataset} further explored real-world electric vehicle (EV) and battery energy storage system (BESS) data, improving battery management in more realistic settings. 
    
    Although modern data-driven approaches have shown promising progress in specific tasks, designing one model for each battery system and management task can be inefficient, requiring significant research and engineering effort.  Approximately 15 thousands works related to battery management were published over the past five years with 4 thousands in 2024 alone and an annual growth over 12\% (see Supplementary Figure 1). Moreover, aggregating multiple models into one battery management system (BMS) will increase computational costs\cite{NS2022-computationalCost}, which is not practical for real-world applications. A more general method for integrating multiple datasets and management tasks is needed but has not yet been developed.

    This work aims to reduce the research effort for developing data-driven battery management algorithms. With the rapid increase in publicly available LiB data \cite{EES2022-dataset-HUST, JPS2011-dataset-CALCE1, JES-dataset-UL-PUR, JPS2021-dataset-RWTH}
    and the growing need for various battery management tasks, it is possible to address the multi-task limitations by extending the success of the pretraining-finetuning paradigm from natural language processing\cite{ JMLR2020-T5}, computer vision\cite{ICLR2021-vit}, and other scientific domains\cite{Nature2024-pathology, Nature2023-material, Nature2024-math, Nature2023-networkbiology} into the field of battery management, allowing for various tasks to be handled using one model with affordable extra costs. 
    
    Unlike vision and language, where downstream tasks have a consistent input format, pre-training on battery data faces a major obstacle due to data format heterogeneity across tasks. Cell-level state estimation focuses on extracting patterns from basic cell data\cite{NC2022-voltageRelaxation-dataset, AE2018-capacity, NC2024-capacity}, whereas life or degradation prediction relies on identifying time-correlated degradation mode across cycles\cite{NE2019-RUL-dataset, NMI2025-RUL-BliNet}. In contrast, system-level anomaly detection must account for cell-to-cell variations\cite{NC2023-dyad-dataset, eTrans2020-fault}. Divergent charging standards also lead to diverse battery data formats. China's mandatory GB/T 32960\cite{GB/T32960.3-2016} requires system-level voltage, while the globally prevalent IEC 61851\cite{IEC61851-24-2023} does not specify. Moreover, real-world LiB systems often allow low-priority data channels to be missed to accelerate system-cloud communications. This data heterogeneity hinders efficient pretraining on battery data, preventing the full utilization of existing pretraining pipelines that require a fixed data format.

    To this end, we introduce a flexible pretrained model, termed Flexible Masked autoencoder (FMAE), which handles diverse data formats across battery management tasks. Our model features carefully designed learnable channel tokens and inter-snippet embeddings to allow pretraining with optional system-level and temporal variations, such as extreme values of individual cells. We compared FMAE with expert-designed features and the state-of-the-art deep learning models across 11 datasets from different LiB systems with varied battery management tasks. FMAE can consistently improve performances simply by fine-tuning the pre-trained model with marginal amount of labeled battery data. 
    
   Our results demonstrate the effectiveness of the pretraining-finetuning paradigm for efficient and accurate real-world battery management.  For example, FMAE achieves an average absolute error of 0.63\% in SOH estimation, improving upon the 1.04\% error achieved by the supervised methods in \cite{NC-2023-SOH-DNNswarm} even with features from domain experts\cite{AE2018-capacity}. In RUL prediction, our model can predict with only 2 cycles of battery data and achieves a similar RUL prediction accuracy compared to the state of the art model BatLiNet\cite{NMI2025-RUL-BliNet}, which utilizes 50 times more battery information (100 cycles). Our work contributes towards the reliable operation of existing LiB systems in real-world setups. 

\begin{figure}[htbp]
            \centering
            \includegraphics[width=1\linewidth]{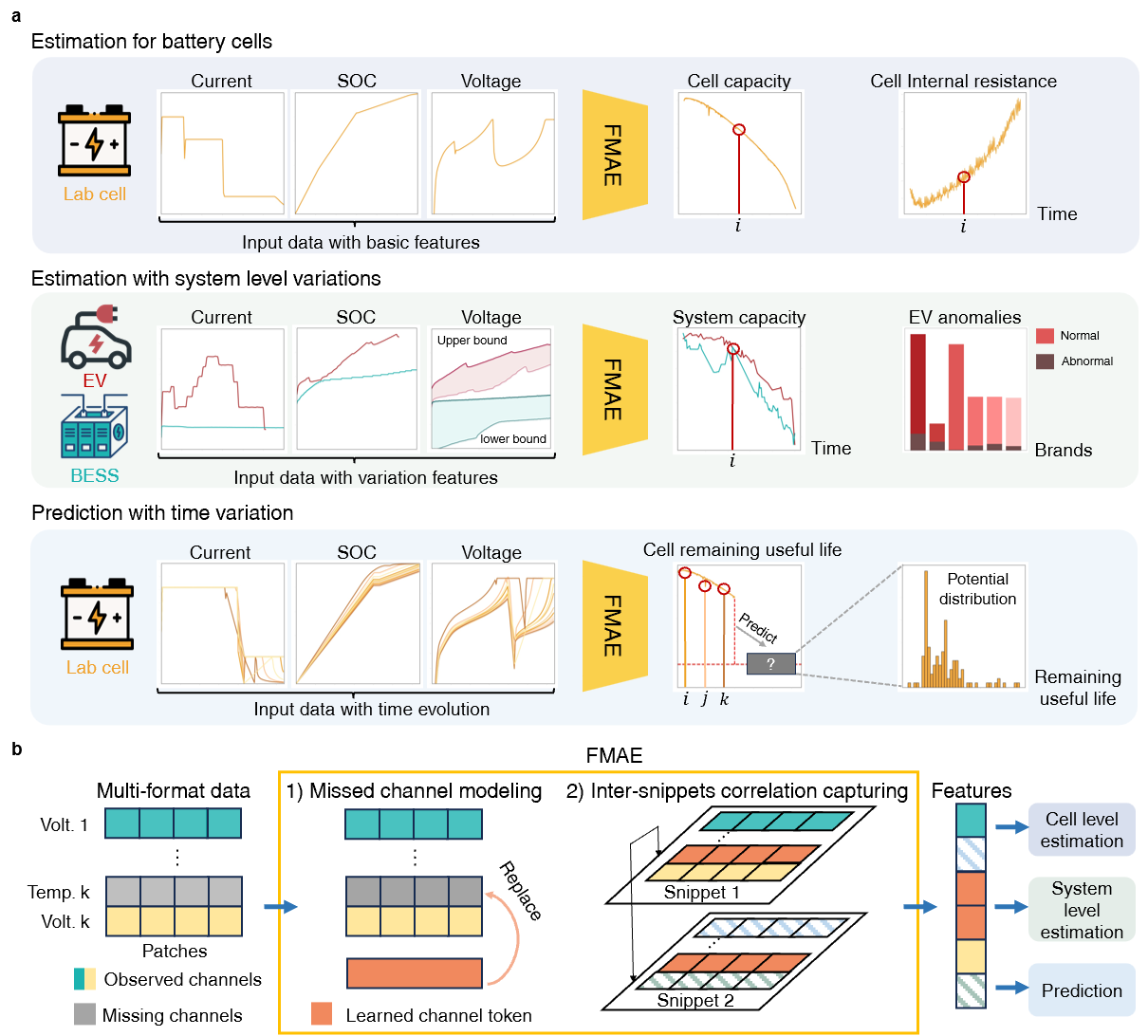}
            \caption{
                \textbf{Overview of the proposed FMAE method and its inference strategy.}
\textbf{a.} FMAE handles diverse data format input without structure modifications. Cell-level estimation utilizes three basic observations (voltage, current, and state of charge) to evaluate its current capacity and internal resistance. System-level estimation evaluates system capacity and detects anomalies by incorporating variations along cells. Prediction infers the remaining useful life of cell through cross-time data snippets.
\textbf{b.} FMAE inference pipeline. With designed pretraining strategy, FMAE can model missed channel data, capture time-correlation knowledge between data snippets, and generalize to different scale battery management tasks.
            }

            \label{fig1}
    \end{figure}
    
\section{Flexible Pretraining for Heterogeneous Battery Data}

To develop a pretrained model for diverse tasks, we compile a dataset (details in Supplementary Table 1) with labels for battery capacity, internal resistance, anomalies, and remaining useful life cycles. 
The data for each task can have diverse format and information and pose immediate challenge for developing a flexible model. In the first row of Figure~\ref{fig1}a, we note that to estimate the capacity and internal resistance of laboratory battery cells, the data typically consists of only basic channels such as current, SOC, and voltage. 
Second, to detect system anomaly, collected data from EV and BESS include additional system-level variations, such as maximum/minimum single voltage and maximum/minimum temperature, as illustrated in the second row of Figure~\ref{fig1}a. These features are critical in health estimation. 
Third, to predict battery states such as the RUL, the model requires multiple charge and discharge information evolved over time, so that it can learn the battery degradation patterns to make predictions (see the third row of Figure~\ref{fig1}a).

\begin{figure}[htbp]
        \centering
        \includegraphics[width=1\linewidth]{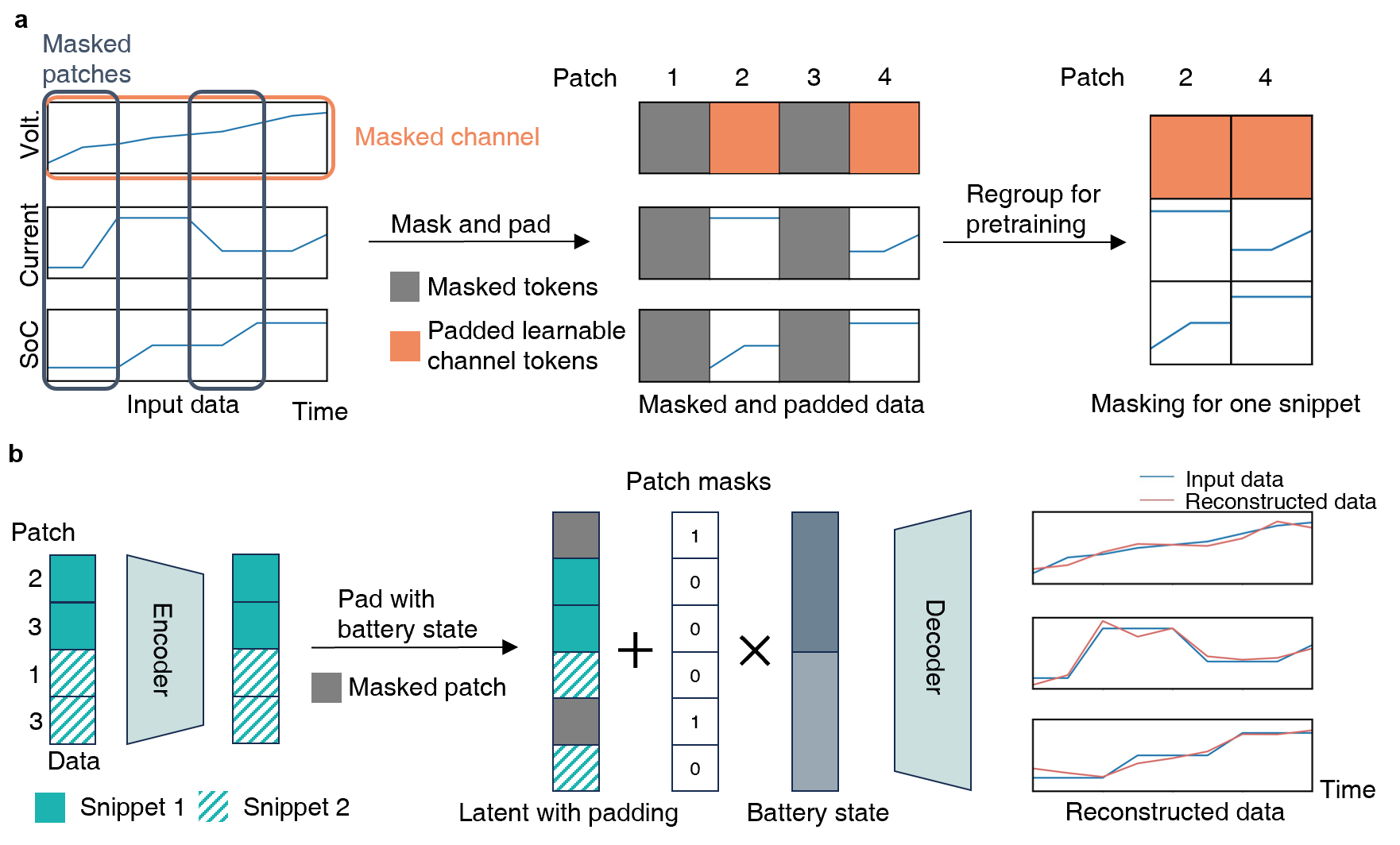}
        \caption{\textbf{The pretraining pipeline of the proposed method with missed channel modeling and inter-snippet correlation capturing.} \textbf{a.} To handle battery data heterogeneity, a portion of the patches and channels in the input data is masked and removed. The masked channels are padded with learnable tokens for maintaining data format during pretraining.  
        \textbf{b.}To further capture inter-snippet correlation, the encoder is applied to the input battery data patches, which is concatenated from multiple masked data snippets. The encoded snippets use the battery state (e.g., the current, SoC, and mileage) to avoid collapsed predictions, and the decoder reconstructs the original input signals from them. 
        }
        \label{fig2}
\end{figure}

Therefore, our method must accommodate different data formats arising from the heterogeneity of battery data, which is incompatible with known pretraining algorithms. To this end, building upon the Masked Autoencoder (MAE) \cite{he2022masked}, we introduce unique learning designs, including missing channel modeling and inter-snippet correlation. We then propose a complete framework from battery pretraining to inference, called FMAE. In addition to randomly masking patches in data as a standard MAE, FMAE randomly masks data channels and battery charging/discharging cycles. Figures~\ref{fig1}b and \ref{fig2} show the FMAE inference and pretraining pipeline, respectively. We note that naively masking battery data would cause model collapse. We highlight our design in the Figure~\ref{fig2} and explain below.

\paragraph{Pretrain with missing channels} To apply our model to laboratory data or broader scenarios with missing channels, FMAE masks not only patches but also channels. During pretraining, FMAE employs learnable tokens to pad the masked channels, while in the inference phase, these tokens serve as the modeling of missing channels, thereby resolving potential channel absence when applied to downstream tasks (see Figure~\ref{fig2}a).

\paragraph{Inter-snippet correlation capturing} 
To reveal signals from time-evolving snippets for prediction, FMAE simultaneously takes multiple masked snippets as inputs. We propose to apply embedded battery states, which are derived from the current, SoC, and mileage of the corresponding snippets (see Figure~\ref{fig2}b), at the decoder. We note that when handling multiple masked snippets simultaneously, simply padding the learnable mask tokens on the positions of the masked patches in the latent space with vanilla position embedding would generate a collapsed model output---the transformer would produce identical outputs for the same positions of masked patches across different snippets. Replacing the vanilla position embedding with embedded battery states not only ensures the proper functioning of the transformer but also, through the incorporation of mileage information, accounts for the effects induced by time variations.

\section{Results}

We collected datasets from 6 EVs, 1 BESS, and 4 laboratory sources (MIT1\cite{NE2019-RUL-dataset}, MIT2\cite{Nature2020-dataset}, KIT\cite{NC2022-voltageRelaxation-dataset}, THU\cite{etran2024-capacity-dataset}) to evaluate the capability of FMAE in handling cell-level features, system-level variances, and temporal variation across diverse LiB systems. A total of 3,792,523 battery data snippets were analyzed, each annotated with multi-task labels for five downstream objectives. Detailed visualizations and descriptions are in Supplementary Figure 2 and Supplementary Table 1. 

We compare FMAE with strong baselines on cell-level estimation, system-level estimation, and RUL prediction tasks. For all these tasks, FMAE achieves improved or comparable performance against previous expert-designed pipelines. FMAE also demonstrates data efficiency by preserving the accuracy even when some data channels or snippets are missing.

    \begin{figure}[h!]
                \centering
                \includegraphics[width=1\linewidth]{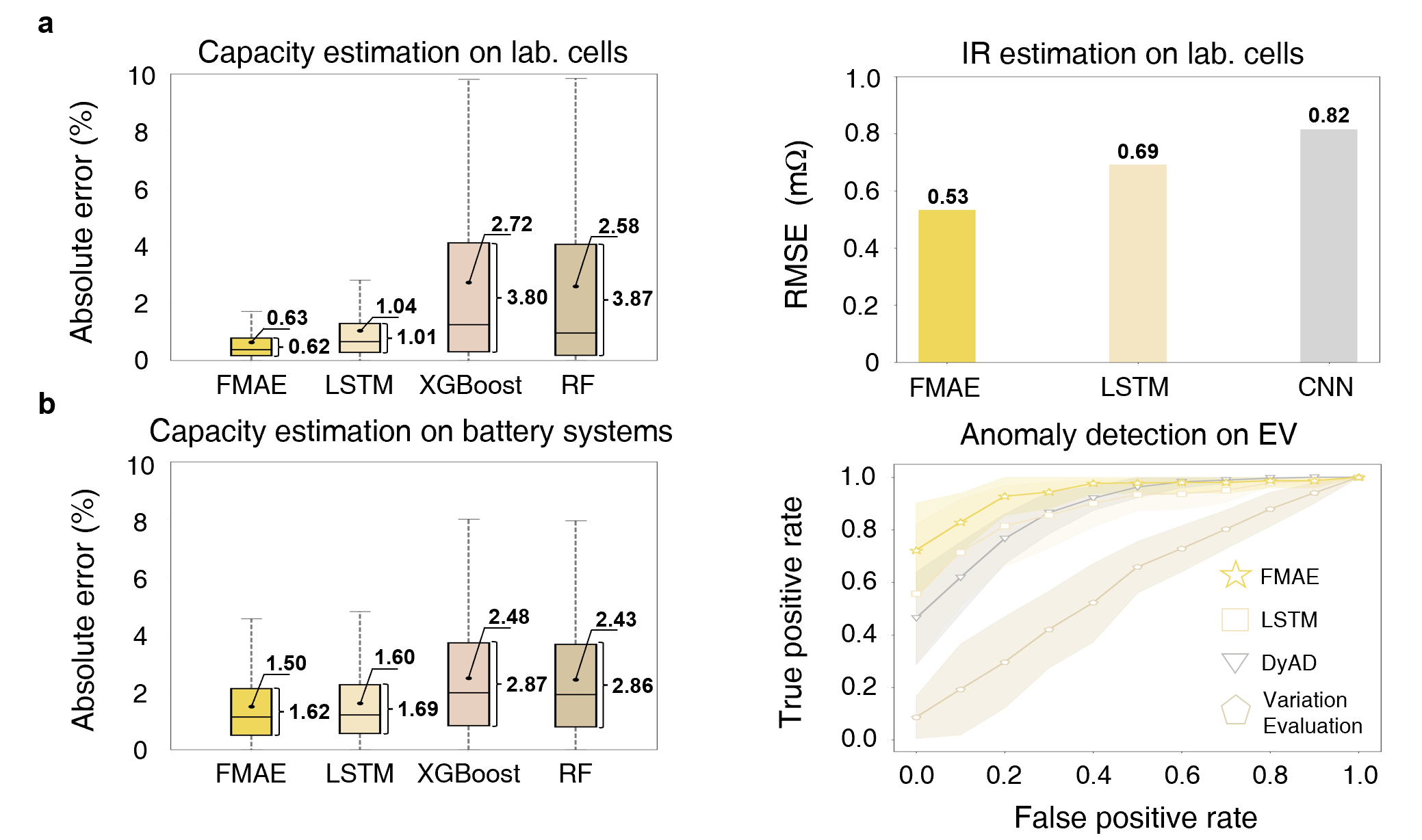}
                \caption{
                \textbf{Comparison of \name{} on cell-level and system-level estimation tasks.}
                \textbf{a.} Cell-level capacity estimation and IR estimation are investigated using two evaluation metrics, absolute error and root mean square error (RMSE). The dotted points in the box plot indicate average SOH errors. The curly braces are the interquartile ranges (IQR) for each method.
                \textbf{b.} System-level capacity estimation and EV anomaly detection use absolute error and the area under receiver operating curve (AUROC) as metrics. The solid curves in the anomaly detection AUROC plot indicate the average values of five-fold cross-validation runs. The shaded regions are the standard deviations of each method.
                }
        \label{fig3}
    \end{figure}
    \paragraph{Cell-level Estimation}
    \label{sec:cell-level}

    We first evaluate FMAE's ability to learn cell features by assessing its performance on well-established cell-level capacity estimation and IR estimation tasks. For cell-level capacity estimation, FMAE significantly outperforms widely adopted baselines and reduces the relative SOH estimation error by 39.4\% to 76.8\%. Specifically, it achieves a mean absolute SOH error of 0.63\%, while the best-tuned baselines even with expert-designed feature~\cite{AE2018-capacity}, 
    achieve higher errors of 1.04\%, 2.72\%, and 2.58\%, respectively. FMAE also demonstrates superior consistency across diverse battery chemistry, as evidenced by its tighter interquartile range of errors, with a value of 0.62, compared to other baselines with values of 1.01 (LSTM), 3.80 (XGBoost), and 3.87 (RF). (Figure~\ref{fig3}a)
    
    For the IR estimation evaluated on the MIT1 dataset, FMAE achieves a root mean square internal resistance error of 0.53 m$\Omega$, surpassing the original supervised neural models\cite{EnAI2021-IR}, where they achieve higher errors of 0.69 m$\Omega$ and 0.82 m$\Omega$, respectively (Figure~\ref{fig3}a). These results collectively demonstrate FMAE's robust capability in modeling essential cell features, providing a foundational basis for the subsequent integration of system-level variations and temporal dynamics from LiB systems.
    
    \paragraph{System-level Estimation}

    Using real-world datasets, we evaluate FMAE's system variation capture capability on system-level capacity estimation and EV anomaly detection tasks. 
    For system-level capacity estimation, FMAE outperforms all baselines previously evaluated at the cell level, achieving a mean absolute SOH error of 1.50\%. This represents relative error reductions of 6.3\% to 39.5\% compared to baselines with higher absolute SOH errors of 1.60\% (LSTM), 2.48\% (XGBoost), and 2.43\% (RF), as shown in Figure~\ref{fig3}b.
    
    We further evaluate FMAE on EV anomaly detection task against the widely studied baselines, including LSTM, dynamical autoencoder(DyAD)\cite{NC2023-dyad-dataset}, and variation evaluation(V-E, a statistical score-based method)\cite{eTrans2020-fault}. Detection performance is quantified by the area under the receiver operating characteristic curve (AUROC), based on the true positive rate and false positive rate. FMAE achieves the best detection results by a 6.7--59.1\% AUROC boost, with an AUROC of 0.945 across studied EV datasets, whereas the rest methods achieve AUROCs of 0.886 (LSTM), 0.885 (DyAD), 0.594 (V-E), as shown in Figure~\ref{fig3}b and Supplementary Table 2. These results demonstrate FMAE's efficiency in capturing system variations across downstream objectives, supporting its real-world applicability in LiB systems.

    \paragraph{Remaining Useful Life Prediction }
    \begin{figure}[h!]
                \centering
                \includegraphics[width=1\linewidth]{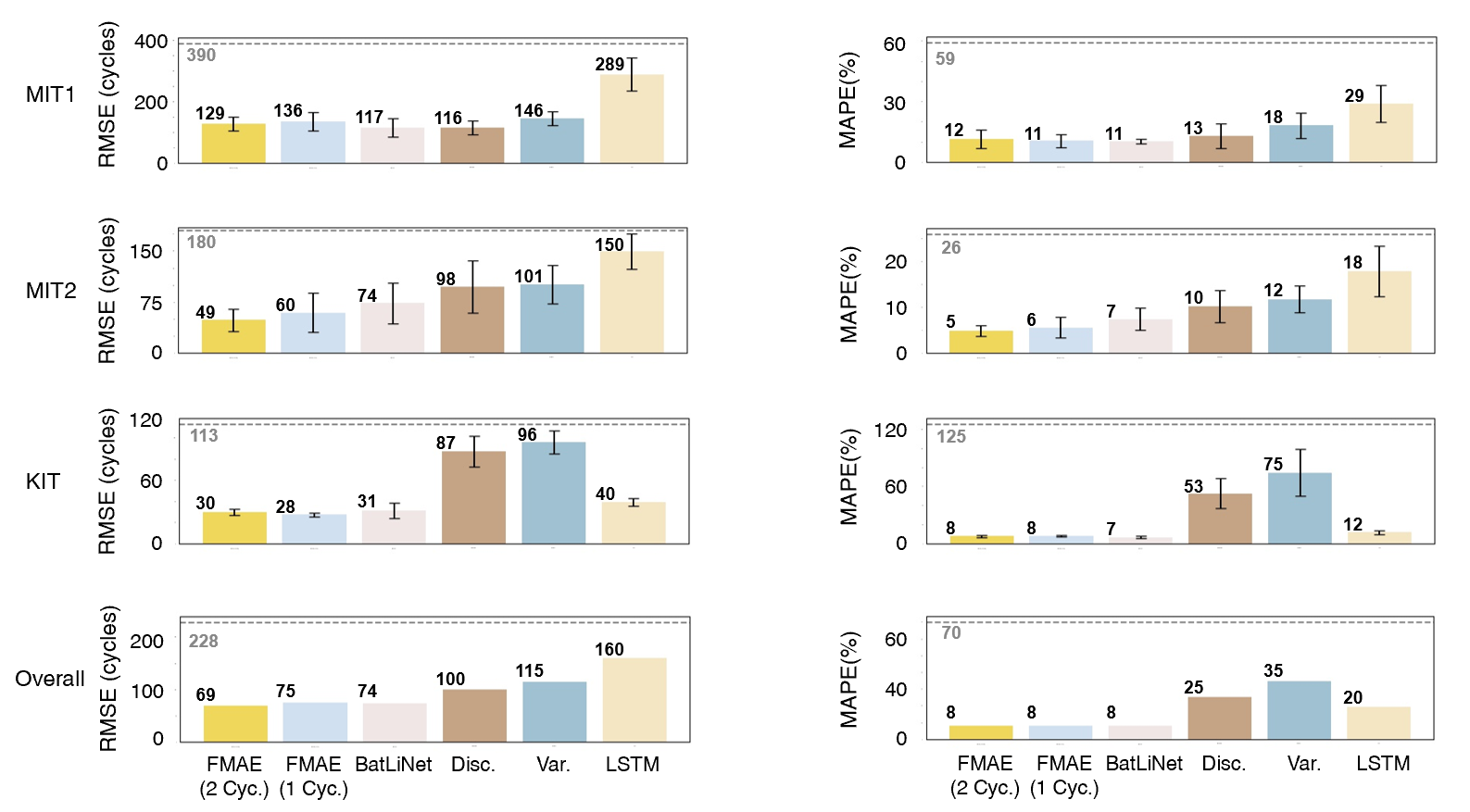}
                \caption{
                \textbf{Comparison of \name{} on RUL prediction task.}
                RUL prediction covers two metrics: root-mean square error (RMSE) and mean absolute point error (MAPE). The grey dotted line indicates a naive baseline, using the average RUL of the training cells to predict the tested cells, demonstrates the complexity of the prediction task. The bar plot shows the mean and standard deviation values for each method.
                We studied linear models, discharge(Disc.) and variance(Var.)\cite{NE2019-RUL-dataset}, deep learning models, BatLiNet\cite{NMI2025-RUL-BliNet}, and the long short term memory(LSTM) for comparison.
                }
        \label{fig4}
    \end{figure}
    Lastly, we experiment \name{}’s ability to predict cells’ RUL, which are affected by various aging factors and complex conditions. 
    In our work, \name{} used fewer battery data and achieved promising results on a par with BatLiNet.
    
    As shown in Figure~\ref{fig4}, with our designed pretraining strategies, FMAE used 2 cycles of input data and achieved 6.8\% and 31.0\% performance boost in RMSE compared with BatLiNet and Disc., respectively, whereas they utilized full 100 cycles of battery information. This result demonstrates the efficiency of our framework on modeling inter-snippet correlations. With limited battery data being observed, FMAE can still recover the unknown system degradation patterns from learned battery knowledge.

\begin{figure}[H]
                \centering
                \includegraphics[width=0.8\linewidth]{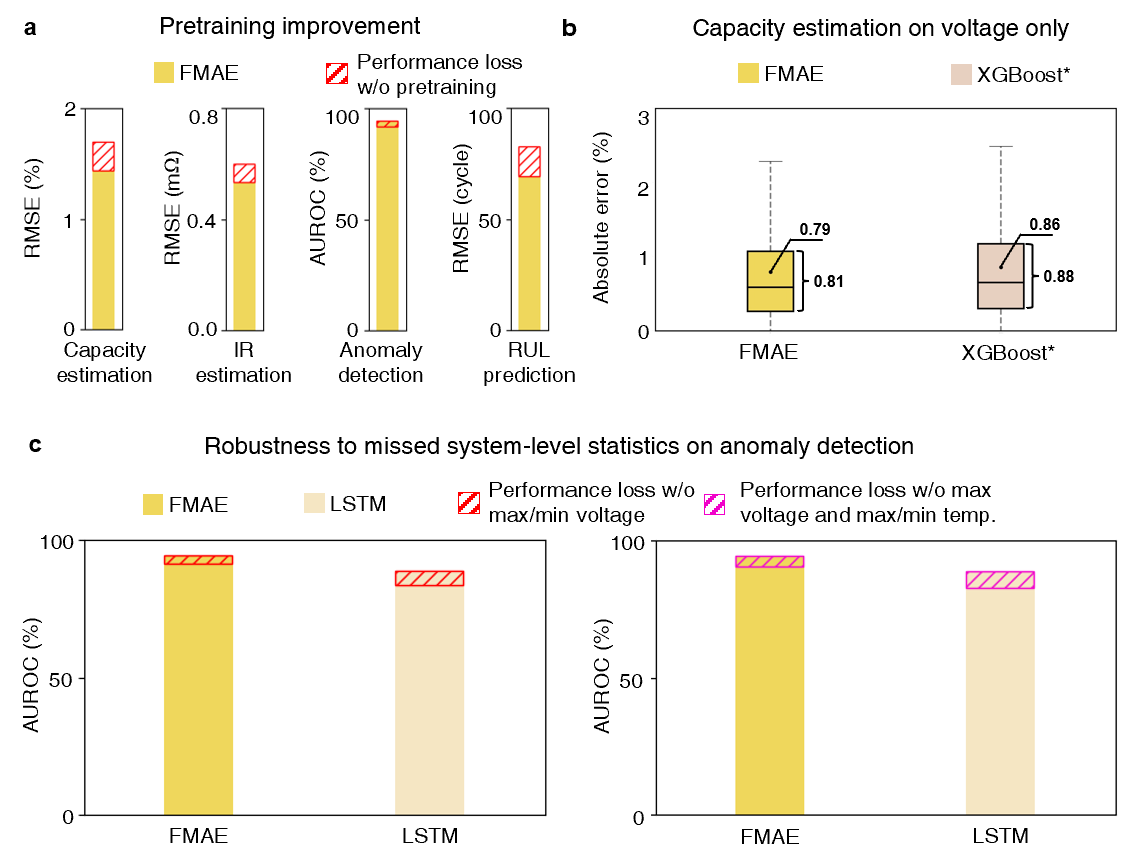}
              \caption{\textbf{Improvements and robustness brought by our pretraining and finetuning strategies.} \textbf{a.} Comparison of results with and without pretraining across four tasks. AUROC is employed as the evaluation metric for anomaly detection, while RMSE is used for the other three tasks. The results of each task are the averaged performance across all datasets available for that task. The hatched area represents the performance loss caused by not using pretraining. \textbf{b.} Comparison of capacity estimation results between FMAE and XGBoost under voltage-only data. $\star$: Features for XGBoost are generated following the work~\cite{NC2022-voltageRelaxation-dataset}. \textbf{c.} Comparison of anomaly detection results between FMAE and LSTM using datasets with artificially partial removal of system-level statistics.}
                \label{fig5}
            \end{figure}

\paragraph{Performance with missing data}
To investigate FMAE's capability in handling potential channel absence, we evaluate its performance under more scenarios. Feature extraction from voltage relaxation \cite{NC2022-voltageRelaxation-dataset} is an effective method for capacity estimation, and we compare FMAE using only a single voltage channel against it. As shown in Figure~\ref{fig5}b, FMAE demonstrates comparable performance, highlighting FMAE’s generalization capability even when deprived of the vast majority of data channels. 

In real-world scenarios, the EVs’ charging and discharging data stored in the cloud from charging piles or vehicle-mounted devices may also lack low-priority data channels, such as volt and temperature. To simulate this scenario, we artificially removed part of the system-level statistics from the EV dataset and detect faulty EVs based on this incomplete data. As shown in Figure~\ref{fig5}c, FMAE employing partial channels still outperforms LSTM utilizing all channels. This result highlights the validity of the proposed missed channel modeling strategy, underscores FMAE's significant potential for real-world deployment.

Last, we compare FMAE with a baseline that trains the same architecture on the same downstream data without model pretraining. As shown in Figure~\ref{fig5}a, FMAE achieves relative performance increases of 15.43\%, 11.02\%, 2.83\% and 16.30\% in capacity estimation, IR estimation, anomaly detection and RUL prediction tasks. This demonstrates that the features learned during pretraining can be broadly and effectively applied, underscoring the significance and effectiveness of pretraining in FMAE.

\section{Discussions}

Our work improves the engineering and data efficiency in reliable management of LiB systems. We present  a flexible pretraining solution, termed FMAE, a novel deep learning framework that equipped with missed channel modeling and inter-snippet correlation capturing strategies, are capable of providing promising results of multitask battery management across 11 different battery chemistry datasets with a single model. We demonstrate that FMAE not only generalize well on existing battery management settings, but also can handle incomplete sensor data, emerge learned knowledge from the pretraining phase, and provide robust inference results when encountered with the real-world missed channel scenarios. 

FMAE has the potential to model more diverse energy system management tasks, such as imputation, forecast, classification, and control, as they also face data heterogeneity challenges along with coupled system interaction and dynamics. FMAE can also be applied to other types of energy systems that share similar data heterogeneity challenges as LiB systems, such as hydrogen energy systems including electrolysers and fuel cells, power network systems, etc. By fully utilizing our framework with energy data, we can capture and understand dependencies across different energy data features/channels and temporal correlations, even when some features are missing or incomplete. Our framework provides a more general and robust solution for modeling and decision-making in future energy systems.

\section*{Methods}
\subsection*{FMAE pretraining}
The FMAE primarily consists of three components: input data masking, an encoder, and a decoder. We provide more details of FMAE pretraining as follows.

    \textbf{Patchify.}
Let $X = [X_1, X_2, \dots, X_n]$ be one input containing $n$ snippets.
Each snippet $X_i \in \mathbb R^{l\times c}$, where $l$ is the length and $c$ is the number of channels.
We follow MAE \cite{he2022masked} and partition each snippet $X_i$ into $s$ patches $[\hat X_{i,1}, \hat X_{i,2}, \dots, \hat X_{i, s}]$, where $\hat X_{i,j}\in \mathbb R^{l_0 \times c}$ is the $j$-th patch of the $i$-th snippet, and $l_0 = l/s$ is the length of one patch.
\vspace{1ex}

\textbf{Channel masking.}
Let $S \subset \{1, 2, ..., c\}$  be the set of masked channels. Specifically, we sample a random subset $S$ of $[c]$ with cardinality $c p_{\text{channel masking}}$, where $p_{\text{channel masking}}$ is the probability of masking a channel.
It contains the indices of channels that will be masked.
Let $v\in \{0, 1\}^{c}$ be a vector where $v_i$ is $1$ if $i\in S$, and $0$ otherwise. 
For a patch $\hat X_{i,j}$, its embedding
\begin{equation*}
        E_{i,j} = \text{Concat}(W\text{vec}(\hat X_{i,j}\text{diag}(v))+b+Cv, \text{pos}_j),
\end{equation*}
where $W\in \mathbb R^{d\times l_0c}$ and $b\in \mathbb R^{d}$ are parameters of the model, $C\in \mathbb R^{d\times c}$ is learnable channel tokens, and $\text{pos}_j\in \mathbb R^{d_{\text{pos}}}$ is the position embedding of $j$, following the sinusoidal positional encoding \cite{vaswani2017attention}.
\vspace{1ex}

\textbf{Patch masking.}
Let $\hat S_1, \hat S_2, \dots, \hat S_n$ be $n$ random subsets of $[s]$ with cardinality $s p_{\text{patch masking}}$, where $p_{\text{patch masking}}$ is the probability of masking a patch.
These $n$ sets contain the indices of patches of $n$ snippets that will be masked.
Let $E_{\text{retain}, i}=[E_{i,j}]_{j\in [s]\backslash \hat S_i}$ and $E_{\text{retain}}=[E_{\text{retain}, 1}, E_{\text{retain}, 2}, \dots, E_{\text{retain}, n}]$.
It denotes the sequence of embeddings to the encoder after discarding the masked patches.
\vspace{1ex}

\textbf{Encoder.}
The encoder architecture follows MAE, and maps the sequence of embeddings $E_{\text{retain}}$ to latent representations $\hat z_{\text{retain}}=[\hat z_{\text{retain}, 1}, \hat z_{\text{retain}, 2}, \dots, \hat z_{\text{retain}, n}]$, where $\hat z_{\text{retain}, i}=[\hat z_{i,j}]_{j\in [s]\backslash \hat S_i}$.
To match the width of decoder, for $i\in [n]$ and $j\in [s]\backslash \hat S_i$, we pass each $\hat z_{i,j}$ through a shared-weight linear layer to obtain $z_{i,j}\in \mathbb R^{d'}$.
\vspace{1ex}

\textbf{Decoder.}
The input to the decoder is not only these latent representations but also the battery states padded into the positions of the masked patches.
Specifically, let $z_i=[z_{i,1}, z_{i,2}, \dots, z_{i,s}]$ for $i\in [n]$ and $z=[z_1, z_2, \dots, z_n]$, where for those $j\in \hat S_i$, $z_{i,j}$ is defined as the result of applying a linear projection to the channels including current, SoC, and mileage of the patch $\hat X_{i,j}$.
After that, a $d'_\text{pos}$-dimensional sinusoidal positional embedding is concatenated to each $z_{i,j}$.
The decoder architecture also follows MAE, and reconstructs the input from the latent representation sequence $z$.
\vspace{1ex}

\textbf{Loss.}
The loss is defined as the mean squared error between the reconstructed results and the input on the masked parts.

The model is trained for 800 epochs on six EV datasets, with the first 40 epochs used for warmup. During the warmup period, the learning rate linearly increases from 0 to 0.00015, followed by a cosine learning rate decay to 0 over the remaining 760 epochs. The Adam optimizer is employed, with each batch containing 256 groups of snippets, where each group consisted of five randomly selected snippets from the same vehicle. Other hyperparameters can be found in the Supplementary Table 3.

\subsection*{FMAE finetuning and inference}
The FMAE is finetuned and inferred separately for each task and each dataset, with all tasks employ five-fold cross-validation. For tasks other than anomaly detection, vehicles/cells in each dataset were directly split into five folds, with four folds used for finetuning and the remaining fold for evaluating model performance. For anomaly detection, normal and faulty vehicles are separately divided into five folds each. During finetuning, four folds of normal vehicles and one fold of faulty vehicles were selected, while evaluating utilized one fold of normal vehicles and four folds of faulty vehicles. Details of datasets are provided in Supplementary Table 1.

To enable FMAE to perform estimation and prediction, the model differs from the pretraining in the following ways during finetuning and inference:
First, the decoder part of the model is removed and replaced by a linear layer that outputs the estimation or prediction. For each input snippet, the model takes the temporal average of the encoder’s output as the feature, which is then fed into the linear layer.
Second, no masking is applied to patches within the input snippets, and the missing channels are padded using the learned channel tokens from pretraining.

FMAE also has subtle differences when applied to anomaly detection and remaining useful life prediction tasks. Since anomaly detection uses vehicle-level anomaly labels, these labels can be assigned to all snippets of the corresponding vehicle during finetuning. During inference, similar to the DyAD’s approach\cite{NC2023-dyad-dataset}, we use the average of the top 10\% logits across all snippets of a vehicle as the vehicle’s anomaly score for detection.
In the RUL prediction task, FMAE simultaneously takes two snippets separated by 20 cycles as input. The encoder extracts features for both snippets, and the difference between these two features is used as the input to the final linear layer for prediction.

\section*{Data availability}
We release all data in Tsinghua Cloud \url{https://cloud.tsinghua.edu.cn/d/fca1245f527d479d82f5/}.

\section*{Code availability}

The code is open source and can be found at GitHub \url{https://github.com/luhong16/FMAE}.

\section*{Acknowledgment}
G.H. acknowledges support by the National Natural Science Foundation of China under Grant 72342004. H.L. and J.Z. acknowledges support by the National Key R\&D Program of China 2024YFA1015800 and Shanghai Qi Zhi Institute Innovation Program. J.C., X.H. and M.O. acknowledge support by the National Natural Science Foundation of China under Grant 52177217.

    	\newpage

	\bibliography{References}

\end{document}


\title{Supplementary Information: \\ Multitask Battery Management with Flexible Pretraining}

    \author{%
      Hong Lu\textsuperscript{1,\dag},~
      Jiali Chen\textsuperscript{2,\dag},~
      Jingzhao Zhang\textsuperscript{1,\dag},~
      Guannan He\textsuperscript{3~*},~
      Xuebing Han\textsuperscript{2~*},~
      Minggao Ouyang\textsuperscript{2~*}\\
    }

    \footnotetext{1~~IIIS, Tsinghua University; \ Shanghai Qi Zhi Institute}
    \footnotetext{2~~State Key Laboratory of Intelligent Green Vehicle and Mobility, School of Vehicle and Mobility, Tsinghua University.}
	\footnotetext{3~~School of Advanced Manufacturing and Robotics,\ National Engineering Laboratory for Big Data Analysis and Applications, Peking University.}
	\footnotetext{\dag~~These authors contributed equally: Hong Lu, Jiali Chen, Jingzhao Zhang, }
	\footnotetext{*~~Email: gnhe@pku.edu.cn, hanxuebing@tsinghua.edu.cn, ouymg@tsinghua.edu.cn}

    \maketitle 
    
    \renewcommand{\figurename}{Supplementary Figure}
    \renewcommand{\refname}{Supplementary References}

    \setcounter{figure}{0}

    \paragraph{Supplementary Figure 1: Publications related to battery management.}
	Supplementary Figure 1 is the trend of global publications we obtained from Scopus using query: TITLE-ABS-KEY ( "lithium-ion battery"  OR  "Battery management" )  AND  TITLE-ABS-KEY ( "State of health"  OR  "State of health estimation"  OR  "battery lifetime prediction"  OR  "battery lifetime"  OR  "anomaly detection"  OR  "Internal resistance estimation"  OR  "state of charge"  OR  "Remaining useful life"  OR  "Remaining useful life prediction"  OR  "Capacity estimation" )  AND  PUBYEAR  $>$  2004  AND  PUBYEAR  $<$  2026, as the time we written the article.
	\begin{figure}[htbp]
    \centering
	\includegraphics[width=1\linewidth]{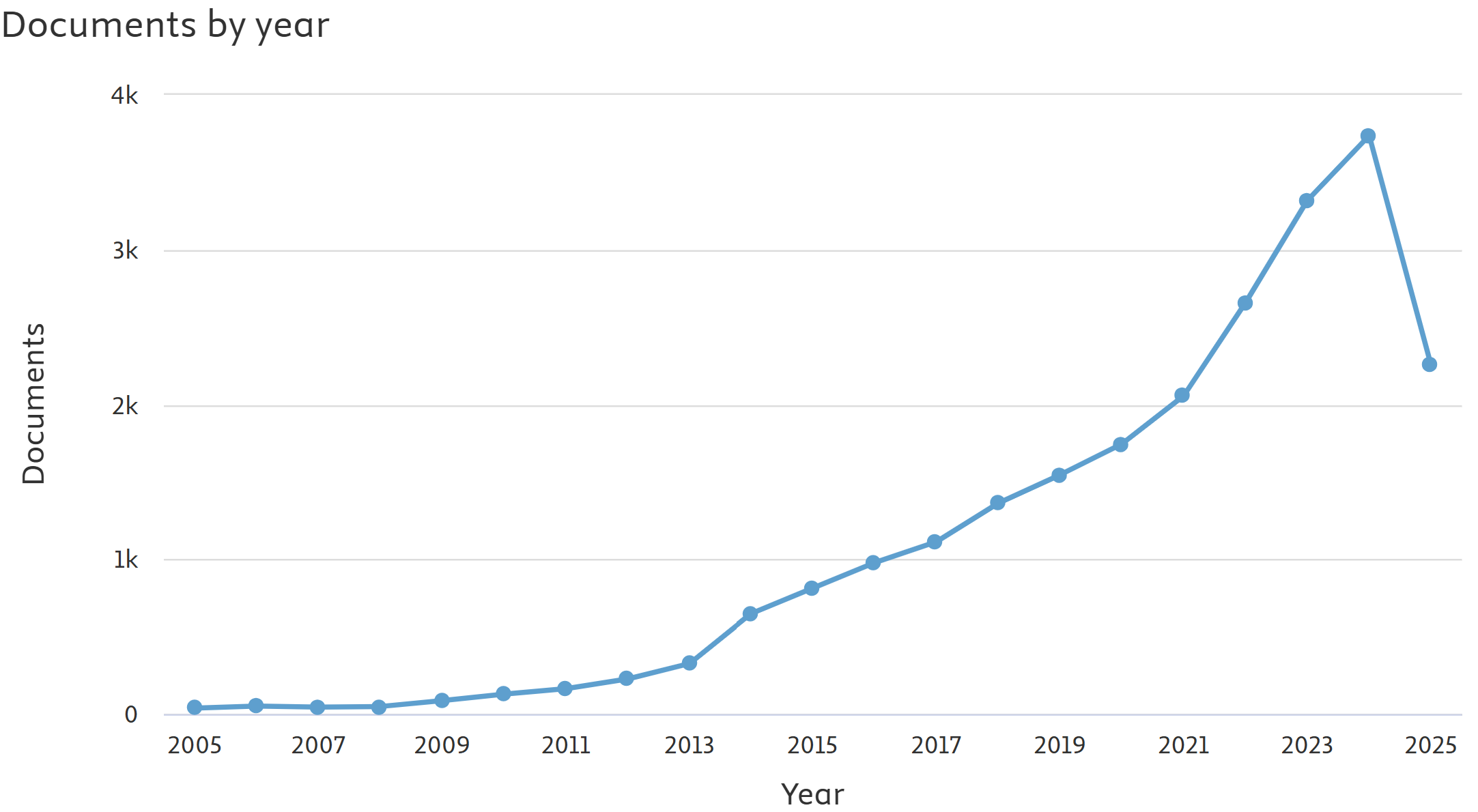}
	\label{SI_fig1} 
\end{figure}
\newpage

\paragraph{Supplementary Figure 2: LiB systems datasets visualization.}
	Supplementary Figure 2 gives an overview of the battery data characteristics from different LiB systems that we studied in the article, where red, yellow, and cyan indicate battery data from EV, Lab, and BESS, respectively. \textbf{a.} The recorded voltage and current of charging snippets from different LiB systems. Voltages from laboratory cells have shorter charging times and irregular charging curves, and this is due to the limited rated capacity of a single cell and the fast-charging protocols inherent in the datasets. The current from EVs is in zig-zag shape and could be caused by the sampling error induced by the sensor or the safety control of battery management systems. The current from BESS is retained in the range between 0.2C -- 0.4C, in which reflects its consistent operational environment compared to battery data from the other two fields. \textbf{b.} The number of abnormal vehicles from different EV datasets. EV anomalies are rare and hard to classify.\cite{NC2023-dyad-dataset} Since the EV3 dataset only contains single-digit abnormal EVs, we eliminated it from the anomaly detection task. \textbf{c.} The RUL distributions of our studied Lab datasets. We eliminated the THU dataset from RUL prediction tasks, since most of the cells did not reach the end of life (80\% of its nominal capacity). \textbf{d.} The normalized capacity degradation curves from all 3 domains. The noisy SOH label of EV datasets is caused by the error accumulation during Ampere-Hour Integration. Part of the accumulation error is induced by entangled charging current between battery pack and vehicle-mounted energy consumption equipment, such as thermal management system, air conditioner, and etc. \textbf{e.} The internal resistance labels from the MIT1 dataset.
	\begin{figure}[htbp]
    \centering
	\includegraphics[width=1\linewidth]{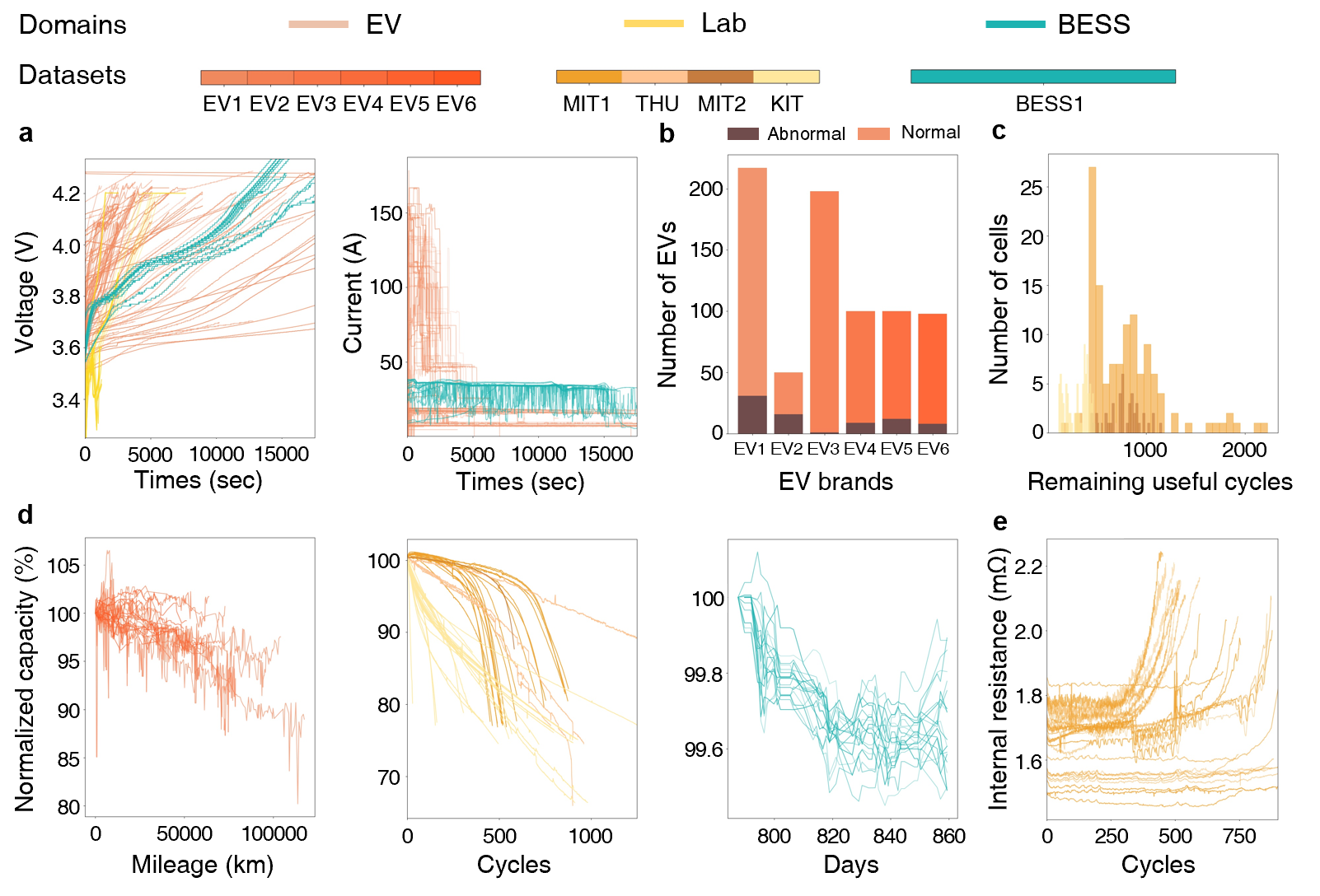}
	\label{SI_fig1} 
    \end{figure}
    \newpage
    
    \paragraph{Supplementary Figure 3: Reconstruction of expert features from battery data snippets}
	Supplementary Figure 3 indicates the effectiveness of our reproduced expert features along with discharge model and variance model from well known baselines\cite{NE2019-RUL-dataset}. We first reconstructed the voltage vs $Q_{100} - Q_{10}$ curve from data snippets, then applied the same feature engineering process and extracted Minimum, Variance, Skewness, Kurtosis along with discharge capacity at cycle 2 and difference between max discharge capacity and discharge capacity at cycle 2, total all six features for model input. Despite using only six features, the discharge model still needs to utilize the full 100 cycles of information to determine the maximum discharge capacity, in order to measure the difference between the maximum discharge capacity and capacity at cycle 2.
	
	\begin{figure}[htbp]
    \centering
	\includegraphics[width=1\linewidth]{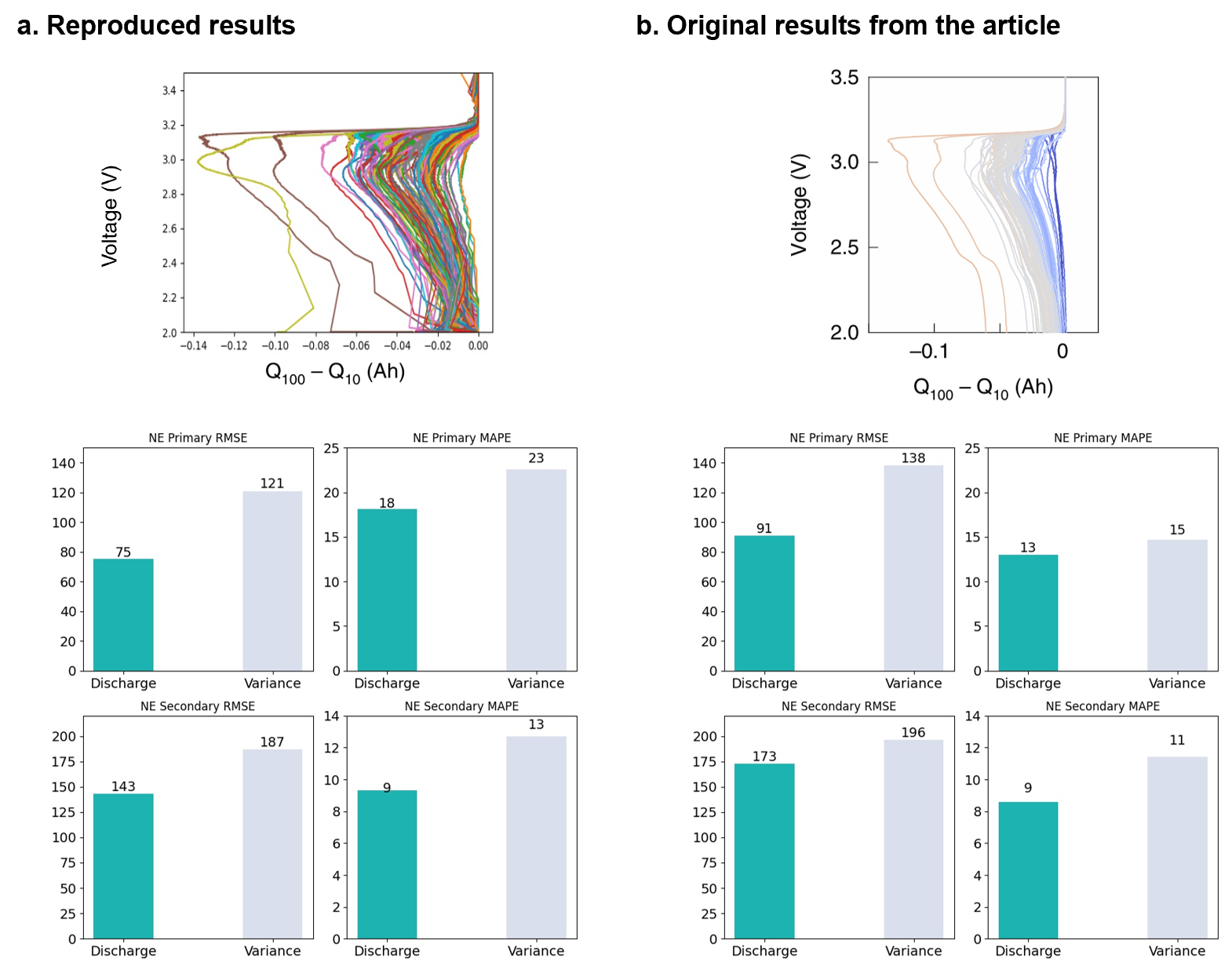}
    \end{figure}
    \newpage

    \paragraph{Supplementary Table 1: LiB systems datasets description and metadata} Supplementary Table 1 gives an overview of how many battery data snippets we used and generated for our experiments, where the dashed line indicates the metadata is unknown due to industries' privacy concerns. We used all eleven datasets with a total of 3,792,523 data snippets. We performed on five major downstream tasks, including cell-level capacity estimation, internal resistance estimation, system-level capacity estimation, EV anomaly detection, and remaining useful life prediction, a total of 20 battery management tasks were studied. The studied LiB systems range from 1.1 to 189 rated capacity in Ah with different battery chemistries.
        \begin{table}[htbp]
    \centering
    \resizebox{\linewidth}{!}{
    \begin{tabular}{lcccccc}
    \toprule
    Dataset & Cell Type & Rated Capacity (Ah) & \# Cells/Packs/Racks & \# Snippets & Tasks\\ 
    \midrule
    \celluline{THU}    & NCM     & 24  & 38  & 66570 & Capacity\\
                           & NCA     & 3.5 & 66  &  \\
    \celluline{KIT}         & NCM     & 3.5 & 55  & 334665 & Capacity, RUL\\        
                           & NCM+NCA & 2.5 & 9   &  \\
    \celluline{MIT1}        & LFP     & 1.1 & 124 & 94225 & Capacity, RUL, IR\\  
    \celluline{MIT2}        & LFP     & 1.1 & 233 & 53741 & Capacity, RUL\\  
    \packuline{EV 1}    & NCM    & 145 & 217 & 1785387 & Anomaly, Capacity\\  
    \packuline{EV 2}    & NCM    & 153 & 50  & 246714 & Anomaly, Capacity\\ 
    \packuline{EV 3}    & NCM    & 169 & 198  & 472829 & Capacity\\ 
    \packuline{EV 4}    & NCM    & -   & 100 & 186765 & Anomaly, Capacity\\ 
    \packuline{EV 5}    & NCM    & -   & 100 & 234664 & Anomaly, Capacity\\ 
    \packuline{EV 6}    & NCM    & -   & 98  & 282689 & Anomaly, Capacity\\ 
    \rackuline{BESS 1}   & NCM-graphite & 189   &7 (17 modules/rack) & 34274 & Capacity\\ 
    Overall & & & & 3792523 &  \\
    \bottomrule
    
    \end{tabular} 
    }
    \resizebox{0.5\linewidth}{!}{
    \begin{tabular}{lcccccc}
    Lab & \textcolor{CategoryCell}{\rule{1cm}{1mm}} &
    EV & \textcolor{CategoryPack}{\rule{1cm}{1mm}} &
    BESS & \textcolor{CategoryRack}{\rule{1cm}{1mm}} \\
    \end{tabular}
    }
    \refstepcounter{table}
        \caption*{}
        \label{SItable1}
    \end{table}

    \newpage

    \paragraph{Supplementary Table 2: Overall results of FMAE compared with baselines across different datasets} The bold font indicates the best result, and the underline indicates the second best. For the anomaly detection task, a higher AUROC means better performance. For capacity estimation, RUL prediction, and IR estimation, a lower RMSE means better performance. FMAE achieved the best overall performance across all battery management tasks. We noticed that some feature based methods achieved the best result on the specific task, which indicates the effectiveness of the specific features that can accurately reveal the underlying patterns in that dataset, and are worth further investigation.
        \label{tab:results}
     \begin{table}[htbp]
        \centering
        \resizebox{\linewidth}{!}{
        \begin{tabular}{c|c|c|c|c|c|c|c|c|c|c|c|c}
    \toprule
    Dataset & EV1 & EV2 & EV3 & EV4 & EV5 & EV6 & BESS1 & THU & MIT1 & MIT2 & KIT & Avg \\
    \midrule
    & \multicolumn{12}{c}{Anomaly detection(AUROC(\%))}
    \\ 
    \hline
    FMAE & \textbf{94.38}    & \textbf{99.76}    & - & \textbf{94.33} & \textbf{98.65}    &\underline{85.19} & - & - & - & - & - & \textbf{94.46} \\
    LSTM & \underline{84.77} & \underline{95.66} & - & \underline{91.25}    & \underline{97.19} & 74.00            & - & - & - & - & - & \underline{88.57} \\
    DyAD\cite{NC2023-dyad-dataset} & 82.26             & 92.49             & - & 85.97             & 87.58             & \textbf{94.05}   & - & - & - & - & - & 88.47 \\
    V-E\cite{eTrans2020-fault}  & 72.26             & 56.36             & - & 44.58             & 56.44             & 67.25            & - & - & - & - & - & 59.38 \\
    \hline
    & \multicolumn{12}{c}{Capacity estimation(RMSE(\%))}\\
    \hline 
    FMAE & \textbf{1.77} & \textbf{1.25} & \textbf{2.49} & \textbf{1.75} & \textbf{1.65} & \underline{1.99} & \textbf{0.97} & \textbf{0.88} & \textbf{0.83} & 0.47 & \textbf{1.73} & \textbf{1.43} \\
    LSTM & \underline{1.81} & \underline{1.37} & \underline{2.77} & \underline{2.10} & \underline{1.81} & \textbf{1.90} & \underline{0.98} & \underline{1.21} & 1.38 & 0.83 & \underline{2.35} & \underline{1.68} \\
    RF\cite{AE2018-capacity} & 3.38 & 1.87 & 3.20 & 2.82 & 2.83 & 2.90 & 1.01 & 4.74 & \underline{0.99} & \textbf{0.27} & 4.99 & 2.64 \\
    XGBoost & 3.39 & 1.90 & 3.26 & 2.89 & 2.84 & 2.93 & 1.02 & 4.82 & 1.03 & \underline{0.41} & 5.05 & 2.69 \\
    \hline
    & \multicolumn{12}{c}{RUL prediction(RMSE(cycles))}\\
    \hline
    FMAE(2 Cycles)  & - & - & - & - & - & - & - & - & 129            & \textbf{49}   & \underline{30}  & \textbf{69} \\
    FMAE(1 Cycle)   & - & - & - & - & - & - & - & - & 136            & \underline{60}& \textbf{28}     & 75  \\
    BatLiNet\cite{NMI2025-RUL-BliNet}        & - & - & - & - & - & - & - & - & \underline{117}& 74            & 31              & \underline{74} \\
    Discharge\cite{NE2019-RUL-dataset} & - & - & - & - & - & - & - & - & \textbf{116}   & 98            & 87              & 100 \\
    Variance\cite{NE2019-RUL-dataset}  & - & - & - & - & - & - & -  & -& 146            & 101           & 96              & 115 \\
    LSTM            & - & - & - & - & - & - & - & - & 289            & 150           & 40              & 160 \\
    Naive baseline            & - & - & - & - & - & - & - & - & 390            & 180           & 113              & 228 \\
    \hline
    & \multicolumn{12}{c}{IR estimation(RMSE(m$\Omega$))}\\
    \hline
    FMAE & - & - & - & - & - & - & - & - & \textbf{0.53} & - & - & \textbf{0.53} \\
    LSTM & - & - & - & - & - & - & - & - & 0.69 & - & - & 0.69 \\
    CNN\cite{EnAI2021-IR} & - & - & - & - & - & - & - & - & 0.81 & - & - & 0.81 \\
    \bottomrule
    \end{tabular}
    }
    \refstepcounter{table}
        \caption*{}
        \label{SItable2}
    
    \end{table}
    \newpage
    
    \paragraph{Supplementary Table 3: Hyperparameters of FMAE pretraining.}  The definitions of variables are provided in the method section.
    \begin{table}[htbp]
        \newcommand{\tabincell}[2]{\begin{tabular}{@{}#1@{}}#2\end{tabular}}
            \centering
            \resizebox{0.4\linewidth}{!}{
                \begin{tabular} {cc}
                \hline
                $n$ & 5\\ 
                $l$ & 128\\ 
                $c$ & 8 \\ 
                $s$ & 8 \\ 
                $p_\text{patch masking}$ & 0.5 \\ 
                $p_\text{channel masking}$ & 0.4 \\ 
                $d$ & 96 \\ 
                $d_{\text{pos}}$ & 12\\ 
                \# layers of the encoder & 6\\ 
                \# heads of the encoder & 3\\ 
                $d'$ & 64 \\ 
                $d'_{\text{pos}}$ & 8\\ 
                \# layers of the decoder & 4\\ 
                \# heads of the decoder & 4\\ %
                \hline
                \end{tabular}
                }
        \refstepcounter{table}
        \caption*{}
        \label{SItable3}
    \end{table} 
    \newpage
    
    \paragraph{Supplementary Table 4: Hyperparameters of baseline methods.} We evaluated baselines with a variety of hyperparameters and reported the best-performing results.
    \begin{table}[htbp]
    \centering
    \resizebox{\linewidth}{!}{
    \begin{tabular}{c|c|c|c}
    \toprule
    Methods & Hyperparameter & Choices & Total methods \\
    \midrule
    RF & Num of trees& [6, 25, 50, 100, 250, 500, 1000] & 35\\
       & Tree depths & [3, 6, 7, 9, 11]\\ 
        \hline 
       & Num of trees& [6, 25, 50, 100, 250, 500, 1000]\\
    XGBoost & Tree depths & [3, 6, 7, 9, 11] & 105\\ 
       & Learning rate & [0.8, 0.5, 0.2] \\
       \hline
    & Conv1d in channels & [3, 24, 32, 32, 64, 64,64]\\
    & Conv1d out channels & [24, 32, 32, 64, 64, 64, 64]\\
    & Conv1d kernel size & [6, 3, 3, 3, 3, 3, 3]\\
    & MaxPool1d kernel size & [2, 2, 2, 2] \\
    CNN & MaxPool1d stride & [2, 2, 2, 2]  & 1\\
    & BatchNorm1d num of features & [24, 32, 64, 64] \\
    & Linear in features & [256, 64] \\
    & Linear out features& [64, 1] \\
    & Dropout rate & 0.4 \\
    \hline
    & V score weights &[0.2286]&\\
    & T score weights &[0.1242, 0.1742, 0.2242, 0.2742, 0.3242, 0.4242]&\\
    V-E& R score weights &[0.1774, 0.2274, 0.2774, 0.3274, 0.3773,0.4774]& 10\\
    & Q score weights &[0.0999]&\\
    & E score weights &[0.0699,0.1699, 0.2699, 0.3699]&\\
    \hline
    & l1 lambda & [0, 0.2, 0.4, 0.6, 1.0]& \\
    Disc.& l2 lambda & [0, 0.2, 0.4, 0.6, 1.0]& 325\\
    & Learning rate & [0.1, 0.09, 0.08, 0.07, 0.06, 0.05, 0.04, 0.03, 0.02, 0.01, 0.009, 0.008, 0.007] & \\
    \hline
    & l1 lambda & [0, 0.2, 0.4, 0.6, 1.0]& \\
    Var.& l2 lambda & [0, 0.2, 0.4, 0.6, 1.0]& 325\\
    & Learning rate & [0.1, 0.09, 0.08, 0.07, 0.06, 0.05, 0.04, 0.03, 0.02, 0.01, 0.009, 0.008, 0.007] & \\
    \bottomrule
    \refstepcounter{table}
    \label{SItable4}
    \end{tabular}
    }
    \end{table}
    \newpage
    \paragraph{Supplementary Table 5: The comparison between BatLiNet with fewer cycles and FMAE.} Supplementary Table \ref{SItable5} shows the result of FMAE trained on the charging data from the 40th, 60th, 80th, and 100th cycles of cells, and BatLiNet trained on the charging and discharging data from the 40th, 50th, 60th, 70th, 80th, 90th, and 100th cycles of cells. Seven cycles are the minimum for using two convolutional layer and two pooling layer without padding, to keep the architecture of BatLiNet. 
    \begin{table}[htbp]
        \newcommand{\tabincell}[2]{\begin{tabular}{@{}#1@{}}#2\end{tabular}}
            \centering
                \begin{tabular} {ccccc}
                \hline
                 & MIT1 & MIT2 & KIT & Avg RMSE \\ 
                \hline
                FMAE & 129            & 49   & 30  & 69 \\
                BatLiNet & 144            & 75   & 32  & 84 \\
                \hline
                \end{tabular}
        \refstepcounter{table}
        \caption*{}
        \label{SItable5}
    \end{table} 
    
\newpage
\section*{Supplementary Note 1: Experiment Setup
}

\paragraph{Cell-level capacity estimation.}
Four datasets (THU, KIT, MIT1, MIT2) were employed. For each cell in the training set, 5\% of its charging snippets were randomly selected for training. The data snippets contain three channels (voltage, current, SoC) and cycle number information.

\paragraph{System-level capacity estimation.}
Seven datasets (EV1-EV6, and BESS1) were employed. For each vehicle in the training set, 5\% of its charging data snippets were randomly selected for training. The data snippets contain seven channels (voltage, current, SoC, maximum/minimum voltage, maximum/minimum temperature) and mileage information.

\paragraph{IR estimation.}
The MIT1 dataset was employed. For each cell in the training set, 20\% of its charging data snippets were randomly selected for training. The data snippets contain three channels (voltage, current, SoC).

\paragraph{Anomaly detection.}
Five datasets (EV1, EV2, EV4, EV5, EV6) were employed, excluding EV3 due to insufficient abnormal vehicles. For each vehicle in the training set, 20\% of its charging and discharging data snippets were randomly selected for training. The data snippets contain seven channels (voltage, current, SoC, maximum/minimum voltage, maximum/minimum temperature) and mileage information.

\paragraph{RUL prediction.}
Three datasets (MIT1, MIT2, KIT) were employed. While different methods exhibit slight variations in data usage, all methods utilized no more than the first 100 cycles of each cell’s charging and discharging data, containing three channels (voltage, current, SoC) and cycle number information.

	\newpage
\section*{Supplementary Note 2: Training details of methods
}

\paragraph{Flexible masked autoencoder (FMAE).} FMAE is pretrained on the six electric vehicle datasets, with the hyperparameters detailed in Supplementary Table \ref{SItable3}. When finetuning and inference, the missed channels of snippets are masked, and we do not mask patches. The model extracts features from the input snippets through the encoder and processes them via a linear layer to generate the estimation/prediction. For the RUL prediction, the features of two snippets are subtracted and fed into the linear layer. The Adam optimizer is used with learning rates of 0.00625, 0.0625 and 0.00625 for capacity estimation on EV, ES and laboratory datasets, respectively, and learning rates of 0.001, 0.000625 and 0.00125 for IR estimation, anomaly detection and RUL prediction, respectively.

\paragraph{Long short-term memory (LSTM).} A two-layer LSTM is trained on input data directly. The Adam optimizer is used with learning rates of 0.375, 0.0625 and 0.00125 for capacity estimation on EV, ES and laboratory datasets, respectively, and learning rates of 0.00125, 0.000125 and 0.00125 for IR estimation, anomaly detection and RUL prediction, respectively.

For the anomaly detection, because the labels are at the vehicle level, FMAE and LSTM adopt the approach proposed by DyAD\cite{NC2023-dyad-dataset}, utilizing the average of the top 10\% logits from snippets to detect anomalies. For the RUL prediction, both methods utilize only the charging data from the 40th, 60th, 80th, and 100th cycles of each cell.

\paragraph{Random forest (RF).} We employed the default implementation of RF from scikit-learn Python library.\cite{JMLR2011-scikit-learn} For feature engineering, we followed the process presented in this work\cite{AE2018-capacity} and extracted the relative charging capacity value $Q_i$ at the specified voltage from input battery data snippets as our input features for capacity estimation tasks. During training, the charging capacity is further normalized to stabilize the training process.
\paragraph{Extreme gradient boosting (XGBoost).} Extreme gradient boosting is a tree-based ensemble method. We adopted the default implementation from XGBoost python library\cite{KDD2016-XGBoost} and performed hyperparameter tuning as shown in the following table. We used the same input features as RF.
\paragraph{Convolutional neural net (CNN).} The convolutional neural net can be applied to both images and 1-d time series data. We follow the same architecture presented in this work\cite{EnAI2021-IR} and adjusted the hyperparameters to better fit our input battery data snippets.
\paragraph{Dynamical autoencoder.} DyAD employs the VAE architecture and partitions data channels into system inputs and system responses. It generates system responses conditioned on system inputs and detects anomalies by reconstruction error. We adopted the code implementation in this work\cite{NC2023-dyad-dataset} and tuned the hyperparameters.
\paragraph{Variation Evaluation method(V-E)} Variation evaluation is an online score based methods\cite{eTrans2020-fault}. It evaluates the consistency of the battery data across cells, and provides scores based on voltage variation, temperature variation, resistance variation, capacity variation, and SOC variation, or so-called electric quantity variation. We followed the same training approach presented in DyAD\cite{NC2023-dyad-dataset}. 
\paragraph{BatLiNet.} Building upon CNN-based intra-cell learning, BatLiNet incorporates inter-cell learning to predict the lifespan difference between two cells, thereby enabling more stable and accurate RUL prediction. We adopted the code implementation in this work\cite{NMI2025-RUL-BliNet} and tuned the hyperparameters. This methodology utilizes charging and discharging data from the 1st to 100th cycles for RUL prediction.
\paragraph{Discharge model (Disc.) and Variance model (Var.).} These two models are linear models with different input features.\cite{NE2019-RUL-dataset} Both equipped with the l1 and l2 loss to alleviate the overfitting problem of linear models caused by the inherent limited model capacity nature. Disc. model utilizes all six features, yet Var. model only uses Variance as the feature. To obtain the comparable results as shown in the original paper, we applied logarithm to the RUL label, thus making the linear mode easier to learn. In the inference phase, we denormalized the predicted results to recover the real RUL.

\newpage
\section*{Supplementary Note 3: Explanation of model collapsed}
We mention that simply padding the learnable mask tokens with vanilla position embedding on the positions of the masked patches would
generate a collapsed model output.
Here we explain them with formulas. For simplicity, only single-head attention is described here. Multi-head attention follows the same mechanism by just treating the heads independently.

Consider an attention layer with inputs $X\in \mathbb R^{l\times d}$, where $l$ is the length and $d$ is the hidden dimension.
The $i$-th row of $X$ indicates an embedded input $x_i$ at position $i$.
The attention layer can be written as:
\begin{equation*}
        \text{output} = \text{softmax}(XQ(XK)^T)XV
\end{equation*}
where $Q, K, V\in \mathbb R^{d\times d}$ are model parameters.
If we use learnable mask tokens with vanilla position embedding, for the same positions across different snippets, they have the same embedded inputs, denoted as $x_i^{(1)}$ and $x_i^{(2)}$ ($x_i^{(1)}=x_i^{(2)}$).
Thus $\text{softmax}(x_i^{(1)}Q(XK)^T)=\text{softmax}(x_i^{(2)}Q(XK)^T)$ and they would produce the same outputs, which prevent the model from reconstructing different signals.
    
    \newpage
 
	\bibliography{References}